# Explaining Deep Learning Models for Structured Data using Layer-Wise Relevance Propagation

Ihsan Ullah, Andre Rios, Vaibhav Gala and Susan Mckeever

*CeADAR Irelands Centre for Applied AI, Technological University Dublin, Dublin, Ireland*

ABSTRACT

Trust and credibility in machine learning models is bolstered by the ability of a model to explain its decisions. While explainability of deep learning models is a well-known challenge, a further challenge is clarity of the explanation itself, which must be interpreted by downstream users. Layer-wise Relevance Propagation (LRP), an established explainability technique developed for deep models in computer vision, provides intuitive human-readable heat maps of input images. We present the novel application of LRP for the first time with structured datasets using a deep neural network (1D-CNN), for Credit Card Fraud detection and Telecom Customer Churn prediction datasets. We show how LRP is more effective than traditional explainability concepts of Local Interpretable Model-agnostic Explanations (LIME) and Shapley Additive Explanations (SHAP) for explainability. This effectiveness is both local to a sample level and holistic over the whole testing set. We also discuss the significant computational time advantage of LRP (1-2s) over LIME (22s) and SHAP (108s), and thus its potential for real time application scenarios. In addition, our validation of LRP has highlighted features for enhancing model performance, thus opening up a new area of research of using XAI as an approach for feature subset selection.

## 1. Introduction

Explainable Artificial Intelligence (XAI) is about opening the "black box" decision making of machine learning (ML) algorithms so that decisions are transparent and understandable. The users of this capability to explain decisions are the data-scientists, end-users, company personnel, regulatory authorities or indeed any stakeholder who has a valid remit to ask questions about the decision making of such systems. As a research area, XAI incorporates a suite of ML techniques that enables human users to understand, appropriately trust, and effectively manage the emerging generation of artificially intelligent partners [15]. Interest in XAI research has been growing along with the capabilities and applications of modern AI systems. As AI makes its way to our daily lives, it becomes increasingly crucial for us to know how underlying opaque AI algorithms work. XAI has the potential to make AI models more trustworthy, compliant, performant, robust, and easier to develop. That can in turn drive business value and widen adoption of AI solutions

A key development in the complexity of AI systems was the introduction of AlexNet deep model [20], a convolutional neural network (CNN) that utilized two Graphical Processing Units (GPUs) for the first time, enabling the training of a model on a very large training dataset whilst achieving state-of-the-art results. With ten hidden layers in the network, AlexNet was a major leap in deep learning (DL), a branch of ML that produces complex multi-layer models that present particular challenges for explainability. Since AlexNet's unveiling in 2012, other factors have boosted the rapid development of DL: availability of big data, cloud computing growth, powerful embedded chips, reduction in the cost of systems with high computational power and memory, and the achievement of higher performance of DL models over traditional approaches. In some application areas, these models achieve better than human level performance, such as object recognition [18, 44], object detection [26, 16], object tracking [23]) or games (e.g. beating AlphaGo champion [39]), predictions [24], forecasting [12, 11] and health [25].

After more than seven years of continuous improvement in AI based DL model, researchers have now started investigating the issues arising from these systems. In addition to ethical concerns such as privacy or robot autonomy, there are other issues at the heart of ML that are critical to handle. For example: potentially biased decisions due to bias in the training data and the model; a system wrongly predicting/classifying an object with high confidence; lack of understanding of how a decision is taken or what input features were important in this decision. As models become more complex, the challenge of explaining model decisions has grown. This can lead to legal complications, such as the lack of adherence to the "right to explanation" under EU General Data Protection Regulation (GDPR) rule [29]. For example, a customer whose loan application has been rejected has the right to know why his/her application was rejected.

Some models are used to make decisions that have life threatening implications, such as the interpretation of potential cancer scans in healthcare. Currently, a doctor is needed as an intermediate user of the system to take the final decision. Other AI scenarios remove the intermediate user. For example, the use of fully autonomous cars would cede full control to the associated AI based driving system. DL models are at the heart of these types of complex systems. Examples such as these emphasise the critical nature of explaining, understanding and therefore controlling the decisions of DL models.

Explainability means different things, depending upon

ihsan.ullah@tudublin.ie (I. Ullah)
ORCID(s):

the user (audience/ stakeholder) of the explanation and the particular concerns they wish to address via an explanation. For example, an end user (customer) may question the individual decision a model has taken about them. A regulatory authority may query whether the model is compliant with gender, ethnicity and equality. An intermediate user, such as the doctor with the diagnostic scan decision, will want to know what features of the input have resulted in a particular decision - and so forth.

**Scope** In this paper, we have four main contributions. First we report recent research work for explaining AI models. We note that there are several survey articles on explainable-AI e.g. Tjoa, E., & Guan, C [41] discuss explainability in health data, Selvaraju et. al [37] & Margret et al. [30] cover image data for object detection and recognition, and [14] discuss financial data/text data. In addition, detailed surveys on XAI as a field are emerging, such as the detailed and comprehensive survey about explainability, interpretability, understandability covered in [5]. Secondly, for the first time we applied an explainability techniques (Layer wise relevance propagation (LRP)) for the explanation of a DL model trained over structured data as input, in this case a 1-dimensional DL model. LRP typically uses image as input, providing intuitive visual explanations on the input image. In our work, we train a one dimensional CNN (1D-CNN) model and applied LRP in order to highlight influential features of the input structure data. This approach enables us to answer questions such as: Which factors are causing customers to churn? Why did this specific customer leave? What aspects of this transaction deem it to be classified as fraudulent? Thirdly, we compare our LRP technique to two common XAI techniques in the field i.e. LIME and SHAP. Finally, we validate the correctness of the LRP explanations (important features) by our approach. This is done by taking the most influential subset of features and using them as input for training classifiers in order to see their performance i.e. to determine whether the new models are achieving equal or better performance on the subset of influence features highlighted in the explanation (testing set) compared to the models trained over the whole set of features.

The paper is organized as follows: Section 2 gives an overview of related work. Section 3 explains the proposed approach that include the datasets used, pre-processing performed, models trained, and finally model explanation details. Then, Section 4 discusses the results achieved with the proposed approach, highlighting important features, as well as, results with the subset of features. Finally, Section 5 gives some future directions and main conclusions of our paper.

## 2. Related Work

Explaining AI systems is not a new research topic but it is still in its early stages. Several survey articles have been published for the domain, including [5, 27, 13, 30, 10, 43]. Of these, [5] is the most recent and complete, summarising all others into one. This survey of XAI includes details about the concepts, taxonomies, and research work up to December 2019 along with opportunities and challenges for new or future researchers in the field of XAI. Arrieta et al. [5] divide the taxonomy of explainable scenarios into two main categories: (1) Transparent ML models that are self-explanatory or answers all or some of the previous questions (e.g. Linear/Logistic regression, Decision trees, K-NN, rule based learning, general additive models) and (2) Post-hoc models, where a new method is proposed to explain the model for explanation of a decision of a shallow or deep model. The post-hoc category is further divided into model-agnostic, which can be applied to all models to extract specific information about the decision and model-specific, which are specific to the model in use e.g. for SVM, or DL models such as CNN.

In contrasts to Arrieta's transparent model view, Mittelstadt et al. [31] give credence to the black box concept by highlighting an alternative point of view about explanation of AI systems and whether AI scientists can gain more through considering broader concepts. In this work, they focus on 'what if questions', and highlight that decisions of a black box system must be justified or allowed to be discussed and questioned. In [35], pressure is enforced on bringing transparency and trust in AI systems by taking care of issues such as the 'Clever Hans' problem [1] - and providing some level of explanation for decisions being made. The authors categorise explanations based on the content (e.g. explaining learned representations, individual predictions, model behaviour, and representative examples) and their methods (e.g. explaining with surrogates, local perturbations, propagation-based approaches, meta-explanations).

Explainability of DLs for structured data is limited. In the majority of cases, traditional ML techniques such as random forest, XGboost, SVM, logistic regression, etc. are used with explainability techniques LIME [34] and SHAP [28]. These methods for explaining predictions from ML algorithms have become well established in the past few years. It is also important to highlight that the majority of the XAI methods, which use DL networks such as CNN, show heatmaps or saliency type visualizations, where the input to the network is images. These techniques are also applied for other types of data apart from images, including text and time series data. The majority of XAI techniques however are not general in terms that they can't be applied to different ML algorithms and/or input types or both. Hence, here we will discuss briefly the explainability of approaches that uses DL techniques in three main categories of input data i.e. images, text, and time series data.

**E-AI in Images:** A well explored area of research in XAI is proposing models (mainly using CNN [46, 47, 22]) that can interpret and classify an input image. When such models are explained, they benefit from the intuitive visual nature of the input. The portion of the image that influenced the model decision can be highlighted, which is relatively easily understood by different type of recipients e.g. end-user (customer) or data-scientists. For example, researchers found Clever Hans [1] type issues in datasets, which are highly

---
[1]https://simple.wikipedia.org/wiki/Clever_Hans

interpretable for this issue [22].

M. D. Zeiler and R. Fergus [45, 46, 47] contributed approaches to understanding mid and high-level features that a network learns as well as visualizing the kernels and feature maps by proposing a deconvenet model to reconstruct strong and soft activations to highlight influences on the given prediction. In [48], a local loss function was utilized with each convolution layer to learn specific features related to object components. These features result in more interpretable feature maps that support explainability. Google's Model Cards tool [30] helps to provide insight on trained image models, providing bench-marked evaluation information in a variety of conditions. The tool helps to answer concerns in explainability, such as the avoidance of bias. Such model cards can be used/considered for every model before deployment.

Ramprasaath et al. [37] proposed a post-hoc method (proposing a new method to explain an existing model for explanation of its decision) that can be applied to several types of CNN models to visualize and explain the decision it provides. The model, termed Grad-CAM, uses a gradient weighted class activation mapping approach in which the gradient targets a class (e.g. cat) and visualises the activations that help in predicting the correct class. A pixel-level visualization has been proposed in the form of a heatmap that shows where the model is focusing on an output map, and thus influenced the model decision.

Recently, Lapuschkin et al. [22] explained the decisions of nonlinear ML model systems for CV and arcade games. They used LRP [3] and Spectral Relevance Analysis (SpRAy) technique and compared both with Fisher Vector based results to detect and highlight Clever Hans issue in famous dataset (PASCAL VOC). The proposed SpRAy uses spectral clustering on the heatmaps generated by LRP to identify typical and atypical behaviour in a semi-automated manner. This is done by learning some specific behaviours (anomalies) in the decisions of a system over a large dataset, unlike the LRP approach which manually analyses every output. These models helps in identifying serious issues in what a model learns e.g. a wrong area/patch of an image to correctly classify the category.

**E-AI in Time Series data:** The analysis and forecasting of time series (TS) information, like any other area that can benefit from AI, needs to incorporate mechanisms that offer transparency and explainability of its results. However, in DL, the use of these mechanisms for time series is not an easy task due to the temporal dependence of the data. For instance, surrogate solutions like LIME [34] or SHAP [28] do not consider a time ordering of the inputs so their use on TS presents clear limitations.

In [38], authors propose a visualisation tool that works with CNN and allows different views and abstraction levels for a problem of prediction over Multivariate TS defined as a classification problem. The solution proposes the use of saliency maps to uncover the hidden nature of DL models applied to TS. This visualisation strategy helps to identify what parts of the input are responsible for a particular prediction. The idea is to compute the influence of the inputs on the inter-mediated layers of the neural network in two steps: input influence and filter influence. The former is the influence of the input in the output of a particular filter and the latter is the influence of the filter on the final output based on the activation patterns. The method considers the use of a clustering stage of filters and optimisation of the input influence, everything with the goal of discovering the main sources of variations and to find similarities between patterns. However, due to clustering to combine the maps it is time consuming and might not be as fast as other techniques such as LRP which work on the pre-computed gradients.

ML tools are widely used in financial institutions. Due to regulatory reasons and ease of explainability, interpretability, and transparency many institutions use traditional approaches such as decision trees, random forest, regression and generative additive models (GAM), at a cost of lower performance. However, there are examples of DL models that have been applied in financial applications e.g. for forecasting prices, stock, risk assessment, insurance. Taking specific model examples, GAMs are relatively easy and transparent to understand and are used for risk assessments in financial applications [6, 8, 40]. The authors in [7] use traditional XGboost and logistic regression (LR), with LR principally used for comparison purposes. After training the model, the Shapley values [28] from the testing set of the companies are calculated. The testing set contains explanatory variables values. They also use a post-processing phase correlation matrix to interpret the predictive output from a good ML model that provides both accuracy and explainability.

Liu et al. [24] proposed a deep model that uses a sparse autoencoder with a 1D-residual convolutional network and a long short term memory network (LSTM) to predict stock prices. The first model was used to reduce the noise and make the data clean for LSTM. This system showed good results for predicting stock prices through price rate of change. In [19], a decision support system from financial disclosures is proposed. It uses a deep LSTM model to predict whether the stock is going up or down. The authors have also focused on answering whether the DL model can give good results on short-term price movement compared to the traditional approach of the bag of words with logistic regression, SVM, etc. The results show that DL based systems, as well as transfer learning and word-embeddings, improve performance compare to naive models. Whilst the performance of these models is not very high, the approach gives a baseline for future research to using DL in financial data. .

In [21], an AI-based stock market prediction model for financial trade called CLEAR-Trade is proposed which is based on CLEAR (Class Enhanced Attentive Response). CLEAR identifies the regions with high importance/activations and their impact on the decision as well as the categories that are highly related to the important activations. The objective is to visualize the decisions for better interpretability. The results on using S&P 500 Stock Index data show that the model can give helpful information about the decision made which can help a company while adopting AI-based systems in answering and handling the regulatory authorities. They model uses a CNN architecture with a convolution layer, leaky ReLu, and Global average pooling layer, followed by the SoftMax layer to classify into two categories i.e. market going up or down is used. The visualization shows that in the correct cases, the model weights the past 4 days of data heavily, whereas in the incorrect cases, it considers older weeks old data as important. Second, in the correct decisions, it considers open, high, and low values for making a decision. Whereas in the incorrect cases, the model considers trade volumes but it is not a strong indicator of correctly predicting the model future. Third, it can show that in the correct cases, the probability or output values are high compare to when the model incorrectly predicts.

**E-AI in Text data:** DL has shown good performance over text data for application areas such as text classification, text generation, Natural Language processing for chat-bots, etc. Similar to vision, financial, and time-series data, several works have been done on text data to explain what and how the text is classified or sentence is generated [2]. A bi-LSTM is used to classify each sentence in five classes of sentiment. An LRP is used to visualize the important word in the sentence. The LRP relevance values are being examined qualitatively and quantitatively. It showed better results than a gradient base approach.

**Summary:** XAI is a highly active area of research in the machine learning domain, with a variety of general and model/data specific approaches in the field and continuing to emerge. We have discussed the most relevant explainability approaches related to images, time-series/financial data, and text. We avoided the explainability of systems with structured data as input for algorithms like random forest, XG-Boost, etc. and those that uses majority of the time techniques like LIME and SHAP. However, the majority of data in day to day life is in fact structured, tabular data. Organizations want to utilize that data with DL approaches but with explainability.

We focused mainly on LRP, a leading XAI technique for DL models that is typically used for images but can be utilized with modifications for other form of inputs and can provides intuitive visual explanations in the form of heatmaps. It has a number of distinct advantages: it provides intuitive visual explanation highlighting relevant input, fast, and has not been tried with 1-D CNN over structured data. It may provide effective at highlighting the important features (input) that contribute most in a decision e.g. customer churn, credit card fraud detection, loan or insurance rejection. 1-D CNN is never or rarely (not in our knowledge to date) used for structured data but we suggest that it can be, with the sliding kernel approach, learning combination of features in the table that as a group contribute to model decisions.

LRP and 1-D CNN have not been applied to structured data model explainability. In this work, our hypothesis is that we can use them for structured data. We are focusing to utilize and enhance the existing XAI techniques for structured data. Also, we want to show and explain the 1D-CNN model over structured data with LRP and comparing the correctness of LRP over SHAP and LIME in terms of their similarity in selecting important features and time complexity. In the next section, we will discuss proposed approach.

## 3. Proposed Approach

Figure 1 shows the proposed approach architecture that works in two phases, and which is applied in turn to each of two datasets. The first phase consists of pre-processing and training a 1D-CNN. The proposed 1D-CNN is trained over structured data. Once the network has been trained, the trained model is used in the second phase where XAI techniques are used to visualize the important features. Our main focus is on using 1D-CNN with LRP. However, we also showed the features selected by Shapley additive explanation (SHAP) and Local interpretable model-agnostic explanations (LIME) for comparison purpose.

### 3.1. First Phase
The following subsection will discuss in detail the datasets used, and the associated model training. i.e. a and T

### 3.1.1. Datasets
Our aim was to create two sample models that reflect common customer-related business problems, and thus interesting explanation scenarios. We selected two scenarios : (1) The prediction of customer churn in the telecoms section and (2) The identification of fraudulent credit card transactions. We used two public structured datasets from the Kaggle website: The Telecom Customer Churn Prediction Dataset (TCCPD) and the Credit Card Fraud Detection Dataset (CCFDD). The telecom churn dataset, TCCPD, is a medium size dataset with meaningful feature names which can be used to give an in-depth explanation of what and why a feature is selected. In the credit card CCFDD data, the features are anonymous but it is a big and highly imbalanced dataset so is appropriate for evaluating the performance of the proposed 1DCNN. For all models, the datasets are divided in 80% for training and 20% for validation of the network. The results shown are based on 5-fold cross validation, using the training split. All data splits are stratified.

**Telecom Customer Churn Prediction Dataset (TCCPD)**
Companies want to be able to predict if a customer is at risk of churn. Retaining an existing customer is better than getting a new customer. There are two types of customer churn - voluntary and involuntary churn. Voluntary churn is of most interest for the company as it is the individual customer's decision to switch to another company or service provider. Understanding the factors/features that are associated with the customer leaving is important. Each record has initially 19 features which when converted from categorical values to non-categorical becomes 28 features associated with the customer, to be used for training a customer churn prediction model. These features have meaningful feature names, allowing us to interpret explanations with domain level judgement. Further details can be found on the Kaggle competi-tion webpage[2].

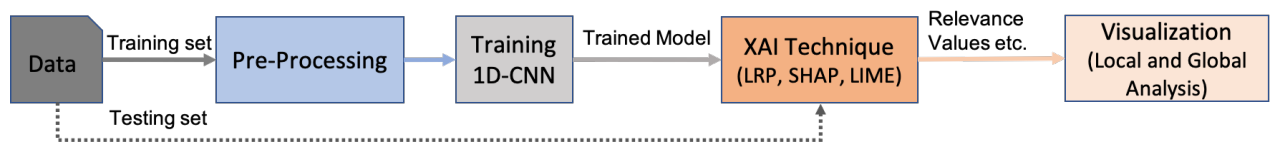

Figure 1: Overall architecture of the proposed approach

**Table 1**
Details of the two datasets. Sample, original and new features are represented by Sam, O-F, and N-F, respectively.

| Name | of Samples | +ve Sam | -ve Sam | %+ve vs %-ve | O-F | N-F |
|---|---|---|---|---|---|---|
| TCCPD | 7043 | 1869 | 5174 | 26.58 vs 73.42% | 19 | 28 |
| CCFDD | 285299 | 492 | 284,807 | 0.172 vs 99.83% | 30 | 30 |

**Credit Card Fraud Detection Dataset (CCFDD)**
Financial companies dealing with credit cards have a vested interest in detecting fraudulent transactions. This highly imbalanced dataset has transactions carried out by European cardholders during September 2013. As shown in Table 1, the dataset is highlighted as imbalanced. Each record contains 30 features out of which 28 are converted by Principal Component Analysis (PCA) and then labelled as V1, V2,...V28. The remaining two features (time and amount) are in their original form. Each record is labelled as either 0 (normal (-ve)) or 1 (fraudulent (+ve)). Further details about this dataset are available at the Kaggle competition webpage[3].

### 3.1.2. Pre-Processing

Our approach uses structured data, consisting of both categorical and numerical data. Data must be numeric in order to use as input to the 1D-CNN network. In addition, the datasets are heavily imbalanced. Our pre-processing steps as follows.
1) Change all the categorical data to numerical data. Where categorical features have more than two values, we create an individual feature for each categorical value, with Yes (1) or No (0) values. The features that were numerical are normalized between 0 and 1 by zero mean and unit variance technique. In TCCPD, there are four features with categorical data. When converted, this results in nine additional features bringing our total dataset features to be 28 (Figure 1).
2) We then apply the SMOTE [9] technique to up-sample the minority class (both datasets) in order to balance the data. SMOTE generates synthetic nearest neighbours for training instances in the minority class in the training set only.
3) Where numeric features are wider than two-choice binary values e.g. monthly charges, those features are first normalized feature-wise, then normalized record-wise in order to obtain a normalized record.

### 3.1.3. Training 1D-CNN

A key innovation in this work is using CNN with structured data and then explaining that deep model. We have selected to use a 1D-CNN for our model. In traditional ML, the process of selecting which features to use for a model is done using various manual steps and domain knowledge (feature engineering). Unlike traditional ML, DL learns important features as part of the training process. We use a 1D-CNN that can slide the kernel across the whole structure to learn important features.

[2]https://www.kaggle.com/pavanraj159/telecom-customer-churn-prediction
[3]https://www.kaggle.com/mlg-ulb/creditcardfraud

**Proposed network structure:**
The proposed base network is a deep seven-layer network that contains 3 convolution layers (with 25, 50, and 100 kernels, respectively), an activation layer after first convolution, two fully connected layer (having 2200 and 2 neurons, respectively), and a SoftMax layer at the end. We used RELU as an activation function. Figure 2 shows the network structure. The kernels in each layer are selected based on the

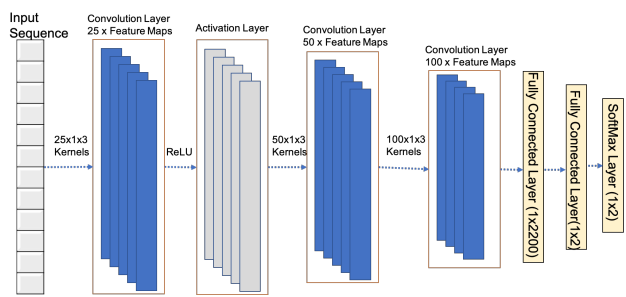

Figure 2: Structure for the proposed baseline 1D-CNN Network

concept of gradual increase or decrease rather than random as being suggested in [42]. Regarding the depth (number of layers), we have used just three, because of the data size limitation. To determine our optimal network set-up, we first tested the network with several different variations of the network structure, with the proposed model selected based on on the metrics of accuracy, precision and f1-score. We used several hyper-parameters for fine tuning the model. We tested the model with learning rate (*lr*) of 0.001, 0.0001 and 0.00001, batch size of 100, 200, and 300, and a maximum iteration of 15000. We used a base *lr* of 0.00001, batch size of 300 (unless specified differently with respective model), for a maximum iteration of 15000. Table 2 shows the various models that we will be using in our experimental work in this paper. In the Model Name column, our proposed name is in the form M-1D-CNN-n-f*. M is a short for model, n represents numbering of 1,2,..5 to show that these are unique models (models are slightly different in hyper-parameters). Whereas, 'f' represents the number of features used as input i.e. 28 and 16. Finally, the '*' represents the model we trained and tested ourselves with similar data that we used

Table 2
Proposed networks structure. Here convolution, fully connected, and output layers are represented by C, F, and O. The number shows number of kernels/neurons in that layer.

| Model Name | LR | BatchSize | Iterations | Network Structure |
|---|---|---|---|---|
| M-1D-CNN-1-28 | 0.00001 | 300 | 15000 | C25 - C50 - C100 - F2200 - O2 |
| M-1D-CNN-2-28 | 0.00001 | 300 | 15000 | C25 - C50 - C100 - F2200 - O2 |
| M-1D-CNN-3-28 | 0.00001 | 200 | 15000 | C25 - C50 - C100 - F2200 - F500 - F10 - O2 |
| M-1D-CNN-4-16 | 0.00001 | 300 | 15000 | C25 - C50 - C100 - F200 - O2 |
| M-1D-CNN-5-16 | 0.00001 | 300 | 15000 | C25 - C50 - C100 - C200 - F1600 - F800 - O2 |
| M-1D-CNN-1-31* | 0.00001 | 300 | 15000 | C25-C50-C100-C200-F4400-O2 |
| M-1D-CNN-2-31* | 0.0001 | 300 | 15000 | C25-C50-C100-C200-F4400-O2 |
| M-1D-CNN-3-31* | 0.0001 | 200 | 15000 | C25-C50-C100-C200-F4400-O2 |

for proposed models 'M-1D-CNN-n-f'.

### 3.2. Second Phase
#### 3.2.1. XAI Technique (LRP, SHAP, and LIME)

In the second phase, once the 1D-CNN is trained, we use that trained deep model as our trial model for explainability. LRP uses the trained model to generate a heatmap based on its relevance values. Our interest in the use of heatmaps is to find and determine what set of features are the most relevant in the prediction of True Positives (1) and True Negatives (0). Furthermore, our objective is to show the important features not only for an individual sample (local analysis) but also for the whole testing set as an overall global pattern learned by the classifier (global analysis).

We have also generated heatmaps from the trained model for comparison with both SHAP and LIME. All the heatmaps are normalized between 0 and 1. The colour scheme reflects the gradient of values from 0 to 1 as shown in Figure 4.4. In the following sub-sections, a brief description of how these techniques work is given.

**Layer-wise Relevance Propagation (LRP):** LRP is one of the main algorithms for the explainability of networks that uses the back-propagation algorithm [3]. LRP explains a classifier's prediction specific to a given data point by attributing 'Relevance Values' ($R_i$) to important components of the input by using the topology of the trained model itself. It is efficiently utilized on images/videos and text where the output predicted value is used to calculate the relevance value for the neurons in the lower layer. The higher the impact of a neuron in the forward pass, the higher its relevance in the backward pass. This relevance calculation follows through to the input where the highly relevant neurons/features are or will have higher values compared to other neurons. As a result, when visualized, the important input neurons can be clearly highlighted based on which the final decision was taken in the output layer. Figure 3 shows the flow of the relevance value calculation. LRP is currently being widely used with CNNs, and to a lesser extent for LSTM in the XAI domain. Improvements in LRP are an active area of research.

**Local Interpretable Model-agnostic Explanations (LIME):** LIME [34] is currently one of the most well-known methods to explain any classification model from a local point of view. This method is considered agnostic because it does not make any assumptions about how the classifier works. It simply builds a surrogate simple model, intrinsically interpretable, such as a linear regression model around each prediction between the input variables and the corresponding outcome variables. The use of a simple model in a local way allows easier interpretation of the behaviour of the classification model in the vicinity of the instance being predicted. LIME attempts to understand the classification model by perturbing the input variables of a data sample and understanding how the predictions change.

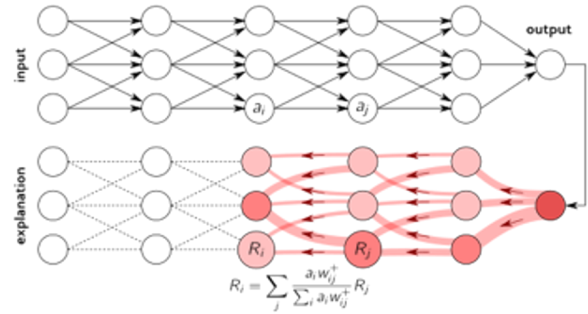

**Figure 3:** Structure for the LRP [3]

In a simplified way, LIME works by first generating random perturbations (fake observations) around the instance to be explained (original instance). Secondly, it calculates a similarity distance between the perturbations and the original instance (e.g. Euclidean distance). Third, it gets the predictions for the perturbations. Followed by picking a set of perturbations with better similarity scores and calculate weights (distance scaled to [0,1]) that represent the importance of these perturbations respect to the original instance. Once perturbations, predictions, and weights have been calculated, it builds a weighted linear regression model, where the coefficients of this simple model will help to explain how changes in the explanatory variables affect the classification outcome for the instance that wishes to be explained. LIME focuses on training local surrogate models to explain individual predictions. Hence, it can be applied to any DL model. However, in terms of time complexity it is simpler than SHAP, but yet require sometime to train.

**Shapley Additive Explanations (SHAP):** SHAP is an approach based on game theory to explain the output of any (but mainly traditional) ML models [28]. It uses Shapley values from game theory to give a different perspective to interpret black-box models by connecting the optimal credit allocation with local explanations and their related extensions. This technique, works as follows: To get the importance of feature $X_i$ it first takes all subsets of features $S$, other than $X_i$. It then computes the effect of the output predictions after adding $X_i$ to all the subsets previously extracted. Finally, it combines all the contributions to compute the marginal contribution of the feature. To avoid recalculation of these subsets, SHAP does not retrain the model with the feature in question left out. It just replaces it with the average value of the feature and generates the predictions. The features which are strongly influential to the model output from the input values are shown. Typically, these influential features are shown in red and other less influential features in blue. This provides a useful clear explanation for simpler models. It is currently (as of this document date) limited to application to traditional ML models due to its time complexity.

### 3.3. Validating the correctness of LRP's highlighted subset of features

We used three XAI techniques for explanation of features. LRP uses a heatmap to highlight the features that have most contributed to the model decision. To validate that the set of features highlighted by XAI techniques (mainly LRP) are genuinely influential on model decisions, we use the highlighted features, with the original dataset labels, to train a simple classifier e.g. Logistic Regression, RF, SVM and see whether the 1DCNN and/or some simple classifier can generate better or equal results with the subset of discriminative features highlighted by XAI techniques. Achieving comparative prediction results will prove that the features highlighted by LRP represent the decision-driving features in the dataset. Rather than manually selecting the highlighted features in the heatmaps, we propose a method that takes account of LRP values for each instance in the dataset, summing to a global ranking for each feature at dataset level. Our approach assumes that all classes are equal weight - i.e. that both labels TP and TN are of equal importance when producing the final feature ranking. The steps of our approach for all three explainability techniques are as follows:

1) Set a threshold value to be used. This threshold represents a cut-off LRP value above which a feature is determined to have contributed to an individual correct test instance.
2) Select all instances that were correctly classified as TP
3) Apply the threshold to the relevance values of each feature of a record in the true positives and true negatives, and converting the feature value to 1 if at or above the threshold, else 0.
4) Sum the rows which have had features converted with 0 or 1, resulting in one vector of dimension n, where n is the number of features.
5) Sort this vector based on the summed values for the features, to produce a numeric ranking of features
6) Repeat step 2-5 for the True Negative (TN) records
7) Select a total of 16 features (top 8 from TP vector and top 8 from TN vector. In case of an odd number, more are selected from TP (one less from TN features))

This is the approach that we have adopted for thresholding and selection of the features. However, more sophisticated techniques, such as allowance for class weighting or merging of LRP values across classes prior to ranking, can be adopted for selecting the features. We also note that there is a lack of strongly negative LRP feature patterns on the global heatmaps. This is the nature of these datasets, and the resultant models. With different datasets/ scenarios, where there is an occurrence of strongly negative LRP feature patterns, the threshold can be adjusted to ensure importance contribution of these features. Next phase optimisation will examine this further. Table 5 and 6 shows an eventual ranked example of the TP case for both datasets. It will be discussed in section 4.

## 4. Results

Every DL network needs to be tuned before one arrives at an optimal model. We trained a similar 1D-CNN network on both datasets. However, some changes were needed in the hyper-parameters (added in following section) to get optimal results. This is mainly because of the size and imbalance nature of the data distribution of the classes. Further, we have worked upon several models for achieving best results and compared the results internally [4] for 1D-CNN model versus several ML classifiers (e.g. Logistic Regression, Random Forest). This is because the dataset was mainly used in a competition on Kaggle and very few papers have used the data.

The following subsections will:
a) Show that 1D-CNN can work on structured data and its performance for both the datasets will be shown in the form of accuracy, precision, recall, specificity, and F1-measure,
b) Discuss the visualized features in a heatmap that play key role in a decision and assessing the results qualitatively (only on telecom churn dataset because the features of CCFDD are anonymous),
c) Compare the highlighted features from the heatmaps of LRP, SHAP, and LIME
d) Validate that the selected subset of features are really important and can show good results when used as input to a simple classifier (done over TCCPD because we know the domain knowledge and the features names).
e) Finally, compare the performance with other techniques.

### 4.1. Performance of 1D-CNN on structured data

The previous highest accuracy on the TCCPD churn dataset was 82.94% [4], achieved with an XGboost classifier having min-max scaler for pre-processing of the 28 features. This result is reported without using cross validation. Looking at Table 3, our models got results less than that as mentioned in above point. However, we got a decent precision of 71.72%. Our highest accuracy with ML classifiers is 79.99% with precision of 64.92% using logistic regression on all the 28 features. Random forest is the classifier which gave 71.72% of precision with 77.77% accuracy. Table 4 shows the results we achieved with proposed model on Credit-card fraud dataset. Our model (M-1D-CNN-1-31) achieved best results in terms of accuracy. This data is highly imbalanced, and due to its complexity, the precision is slightly low in comparison to [17]. However, its overall accuracy is less than ours by a huge margin. Our model also shows better results compared to others reported in [1, 36].

---
[4]Trained by ourselves

### 4.2. Visualizing local and global heatmaps of features using LRP

The main objective of XAI in this work is to see how or why a deep network gave a specific decision (TP, TN, FP, or FN). Figure 4.4 (a) shows a heatmap generated by LRP for local interpretation of a single record that resulted in TP. The heatmap clearly shows that because of features such as the customer being a senior citizen (SeniorCitizen), monthly contract (contract_M_to_M), Fiber optic internet (IS_Fiber_Optic), high monthly charges (MonthlyCharges), and no phone service (PhoneService) it is predicted that this customer is going to churn. Based on domain knowledge, a customer is more likely to churn if the contract is monthly, having no phone service, and have high monthly charges. To retain the customer, it is suggested for the company to reduce some charges by reducing some features but offering yearly or biyearly contract so that to retain the customer.

Similarly, Figure 4.4 (b) shows local analysis of a record which will not churn. The heatmap shows the features for a customer who will not churn are not a senior citizen, have online security, contract of one year without fiber optic, with several other features but less monthly charges. Figure 4.4 (c) shows a clear pattern for all the correctly classified TP samples in the testing set, highlighting the important features that play key role in the model decision towards being marked as TP. Same is the case for TNs as shown in Figure 4.4 (d).

This ability to understand model decisions at class level has a tangible business use case. In our TCCPD example, understanding TP and TN can help a company/data scientist in increasing the revenue by retaining the customers - supporting the well-known maxim that in business, it is easier to keep a customer than to find a new one. On the other hand understanding and recognising the key features for local and global analysis of FP and FN samples may help data-scientists to avoid or reduce the discrepancy in the data results in mis-classifications by the model

### 4.3. Soft comparison of Heatmaps from LRP, SHAP, and LIME

One of the main ideas of our work was to show that LRP can perform well for explainability of the deep model in various forms. SHAP and LIME are two other common techniques used for explainability. Using LRP as the baseline explainability technique, we demonstrate advantages of LRP because it highlights the same features as important/discriminative as those of SHAP and LIME but in far less execution time.

Table 5 and 6 shows the features ranked in descending order for LRP, SHAP, and LIME from the model trained on TCCPD and CCFDD, respectively. The ranking is done based on the approach explained in section 3.3. The value 28 shows that the feature is highly discriminative and is considered important whereas 1 means having low importance in the tables. The features are sorted in descending order of LRP ranking, whilst showing the other rankings of LIME and SHAP with respect to the LRP ranking. As we have taken top and bottom eight features from TP and TN each, therefore, if we consider the important (top) features in the first row of Table 5, and similarly for correspond SHAP and LIME, we can see that the features ranked high (e.g. with 28, 27) in LRP are majority ranked high in SHAP and LIME as well (e.g. Contract_M_M, PhoneService, Tenure, IS_Fiber_Optic, Monthly charges). Five out of eight features are common for all three. In addition, the time taken by LRP running on CPU (MacBook Pro) is 1-2 seconds to generate a heatmap for a single record at test time which is far lower than the time taken by LIME (22 seconds) and SHAP (108 seconds). This shows that LRP is faster. We also show in the section that it selects a highly discriminative feature set. that if used with a simple classifier will generate similar or better performance. This is possible because the ambiguous or redundant features are removed which were confusing the system. The SHAP and LIME code can be slightly optimized by either changing some of its parameters (for example the size of the neighbourhood (we used default of 10, if we increase it increases the time it takes)), parameters for regularisation) however, still it takes more time compared to LRP.

### 4.4. Validation of the subset of LRP highlighted features importance

It is important to verify that the features highlighted as important by LRP are genuinely pivotal in driving the model decisions. Therefore, we use the subset of important features highlighted by LRP as input to traditional ML techniques (e.g. logistic regression, random forest) and proposed 1D-CNN. Table 3 last five rows shows that using features generated from LRP values did not yields good results with traditional ML classifiers. However, we achieved good results with the deep 1D-CNN's (M-1D-CNN-4-16* and M-1D-CNN-5-16*) as shown in Table 3. The interesting point is that our 1D-CNN for LRP surpassed the existing kernel [36] results both in terms of accuracy and precision while trained on this subset of features. We achieved a highest accuracy of 85.54% with precision of 73.995% and f1-score 72.603%. Moreover, we achieved higher f1-score of 73.23% from model with using LRP values as features using SMOTE for balancing the data and with batch size 200 and LR of 0.00001. A key point to note is that using features derived from LRP values, all 1DCNN models gave results close to 83%. The used subset of features contain 16 features. These are selected from the LRP values generated by the process as explained in section 3.3. The good performance with this subset of features proves that the highlighted features are important that can generate almost similar result through a simple classifier instead of a DL model. Hence, if needed in a situation where memory and processing is an issue e.g. when deployed in IoT or on edge device, a smaller simple classifier

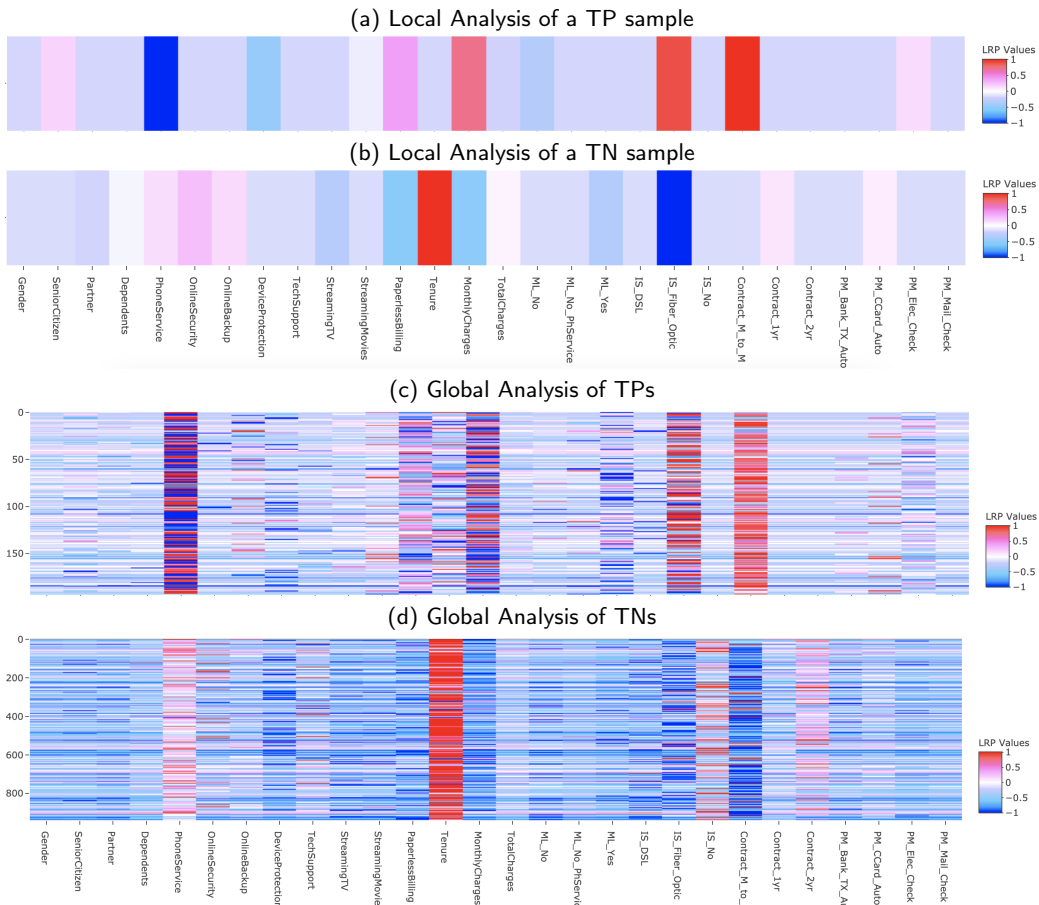

**Figure 4:** Visualizing LRP heatmaps for Local (individual (TP (a) and TN (b))) and global (all TP (c) and TN (d)) samples in Telecom Churn testing set. Feature sequence of (a) and (c) is similar to (b) and (d)

can be used rather than a deep neural network.

### 4.5. Comparison with State-of-the-art

Table 3 shows a detailed comparison of the results achieved by us from our proposed model. In addition, it also contain the results from traditional ML trained by other researchers in research articles and on the Kaggle competition webpage. Many of the results were not based on 5-fold cross validation, therefore, we retrained same techniques to get results with 5-fold cross validation for a fair comparison. One point to highlight is that the published state-of-the-art is only available for the actual original model classification performance i.e. We've no state-of-the-art results for the correctness of features highlighted by LRP.

The results showed that our results from 1D-CNN (M-1D-CNN-1-28*) in the case of using all 28 features is lower than XG-Boost [4] by 0.0003. However, for the same XG Boost-28* when we trained and calculated performance after 5-cross validation, it showed 0.0615 fewer performance than our best model (M-1D-CNN-1-28*) with 28 features as input.

In terms of precision, Random Forest-28* achieved highest precision of 0.7172 that is 0.0478 and 0.0161 higher than Random Forest results reported in [33] and our M-1D-CNN-1-28* model, respectively. However, in terms of F1-score, our model shows better result than that of Random Forest-28* and reported by [33] by 0.2514 and 0.088, respectively.

The state-of-the-art results are achieved when we use our proposed model for selecting subset of features and then using those selected features as input to same networks (1D-CNN and traditional ML techniques) to train and test. Our model M-1D-CNN-5-16* achieved 0.8554 accuracy, 0.7399 precision, and 0.7260 F1-score that are higher than all other models at a good margin. This shows that XAI as an approach for subset selection of discriminative features can give us almost equal or better results with both proposed 1D-CNN model and traditional ML techniques. This can be used as a strategy of first using DL, and then when we have the reduced feature set, using those with a simple classifier which paves the way to investigating this approach for use on embedded or edge devices where there are limitations on memory.

### 5. Conclusions

We have provided the first application of 1D-CNN and LRP on structured data. In terms of accuracy, precision, and F1-Score performance, our deep network performs marginally below the benchmark methodology reported in state-of-the-art on the same data by a small fraction but achieved higher when we used cross-validation on the same model. However, more importantly, we took the initiative for using 1D-CNN+LRP on structured data. Using the approach of 1D-CNN+LRP for validating the subset of features highlighted by LRP as important, the reduced feature set used to train a model can give state-of-the-art performance. Hence, we initiated a new area of research for XAI as a tool for feature subset selection. The proposed approach enhances performance in terms of accuracy, precision, f1 score, and computation time. It also substantially reduced the number of features to be used in a deployed system where resources are limited e.g. an edge device. For future, we plan to do further analysis on other datasets.

Table 3
Comparison of ours (with *) vs. other models on TCCDD. Accuracy, Precision, Specificity, and Cross-validation are represented by Acc, Preci, Speci, and Cross-V, respectively

| Model Name | Train/Test | Acc | Preci | Recall | Speci | F1score | Cross-V |
|---|---|---|---|---|---|---|---|
| **M-1D-CNN-1-28*** | **80/20** | **0.8264** | **0.7011** | **0.6189** | **0.9028** | **0.657** | **Yes** |
| M-1D-CNN-2-28* | 80/20 | 0.8121 | 0.6871 | 0.5556 | 0.9067 | 0.6138 | Yes |
| M-1D-CNN-3-28* | 80/20 | 0.8041 | 0.6727 | 0.5497 | 0.8985 | 0.5978 | Yes |
| Logistic Regression-28* | 80/20 | 0.7993 | 0.6616 | 0.5221 | 0.9015 | 0.583 | Yes |
| Decision Tree-28* | 80/20 | 0.7864 | 0.6875 | 0.3794 | 0.9363 | 0.4881 | Yes |
| Random Forest-28* | 80/20 | 0.7773 | 0.7172 | 0.2842 | 0.9588 | 0.4066 | Yes |
| SVM Linear-28* | 80/20 | 0.7973 | 0.656 | 0.5209 | 0.8991 | 0.5802 | Yes |
| SVM RBF-28* | 80/20 | 0.7785 | 0.6358 | 0.4159 | 0.9122 | 0.5023 | Yes |
| XG Boost-28* | 80/20 | 0.7649 | 0.5727 | 0.4956 | 0.8641 | 0.5311 | Yes |
| **Results in Literature** | – | – | – | – | – | – | – |
| Logistic Regression [33] | 75/25 | 0.8003 | 0.6796 | 0.5367 | – | 0.5998 | No |
| Random Forest [33] | 75/25 | 0.7975 | 0.6694 | 0.4796 | – | 0.569 | No |
| SVM RBF [33] | 75/25 | 0.7622 | 0.5837 | 0.5122 | – | 0.5457 | No |
| Logistic Regression [4] | 70/30 | 0.8075 | – | – | – | – | No |
| Random Forest [4] | 80/20 | 0.8088 | | – | – | – | No |
| SVM | 80/20 | 0.8201 | – | – | – | – | No |
| ADA Boost [4] | 80/20 | 0.8153 | – | – | – | – | No |
| XG Boost-28 [4] | 80/20 | 0.8294 | – | – | – | – | No |
| Logistic Regression [32] | 80/20 | 0.8005 | – | – | – | – | No |
| **Reduced Features Results** | – | – | – | – | – | – | – |
| M-1D-CNN-4-16* | 80/20 | 0.8462 | 0.7339 | 0.6718 | 0.9104 | 0.7014 | Yes |
| **M-1D-CNN-5-16*** | **80/20** | **0.8554** | **0.7399** | **0.713** | **0.908** | **0.726** | **Yes** |
| Logistic Regression-16* | 80/20 | 0.7996 | 0.6633 | 0.52 | 0.9026 | 0.5823 | Yes |
| Decision Tree-16* | 80/20 | 0.7871 | 0.6911 | 0.3788 | 0.9374 | 0.4886 | Yes |
| Random Forest-16* | 80/20 | 0.7807 | 0.7058 | 0.3139 | 0.9521 | 0.4337 | Yes |
| SVM-16* | 80/20 | 0.7986 | 0.6571 | 0.5278 | 0.8984 | 0.5849 | Yes |
| XG Boost-16* | 80/20 | 0.7618 | 0.5645 | 0.5035 | 0.8569 | 0.532 | Yes |

Table 4
Comparison of our 1DCNN (with *) performance on CCFDD against published results

| Model Name | Train/Test | Acc | Prec | Recall | Spec | F1-score | Cross-V |
|---|---|---|---|---|---|---|---|
| **M-1D-CNN-1-31*** | **80/20** | **0.9991** | **0.6732** | **0.9520** | **0.9992** | **0.7866** | **Yes** |
| M-1D-CNN-2-31* | 80/20 | 0.9989 | 0.6507 | 0.8779 | 0.9992 | 0.7446 | Yes |
| M-1D-CNN-3-31* | 80/20 | 0.9987 | 0.6079 | 0.8553 | 0.9990 | 0.7037 | Yes |
| Logistic Regression-31* | 80/20 | 0.9229 | 0.0201 | 0.9044 | 0.9230 | 0.0393 | Yes |
| Decision Tree-31* | 80/20 | 0.9651 | 0.0442 | 0.8762 | 0.9652 | 0.0840 | Yes |
| Random Forest-31* | 80/20 | 0.9946 | 0.2232 | 0.8579 | 0.9948 | 0.3533 | Yes |
| Gaussian NB-31* | 80/20 | 0.9748 | 0.0550 | 0.8474 | 0.9751 | 0.1033 | Yes |
| Logistic Regression [17] | 80/20 | 0.81 | 0.76 | 0.9 | – | 0.82 | No |
| Isolation Forest [36] | 70/30 | 0.997 | – | – | – | 0.63 | No |
| Local Outlier Forest [36] | 70/30 | 0.996 | – | – | – | 0.51 | No |
| SVM [36] | 70/30 | 0.7009 | – | – | – | 0.41 | No |

Table 5
Ranked feature comparison for LRP, SHAP, and LIME over Telecom Churn dataset.

| | Contract_M_to_M | PhoneService | IS_Fiber_Optic | MonthlyCharges | PaperlessBilling | Tenure | PM_CCard_Auto | OnlineBackup | StreamingMovies | ML_Yes | PM_Elec_Check | ML_No_PhService | ML_No | PM_Bank_TX_Auto | StreamingTV | TechSupport | DeviceProtection | Dependents | OnlineSecurity | Partner | SeniorCitizen | PM_Mail_Check | TotalCharges | IS_DSL | IS_No | Contract_1yr | Contract_2yr | Gender |
|---|---|---|---|---|---|---|---|---|---|---|---|---|---|---|---|---|---|---|---|---|---|---|---|---|---|---|---|---|
| LRP | 28 | 27 | 26 | 25 | 24 | 23 | 22 | 21 | 20 | 19 | 18 | 17 | 16 | 12 | 12 | 12 | 12 | 1 | 1 | 1 | 1 | 1 | 1 | 1 | 1 | 1 | 1 | 1 |
| LIME | 26 | 22 | 25 | 1 | 1 | 24 | 1 | 1 | 1 | 1 | 1 | 1 | 1 | 1 | 1 | 21 | 1 | 1 | 23 | 1 | 1 | 1 | 1 | 1 | 26 | 1 | 26 | 1 |
| SHAP | 26 | 25 | 28 | 23 | 1 | 27 | 1 | 1 | 1 | 1 | 1 | 1 | 1 | 1 | 1 | 1 | 1 | 1 | 1 | 1 | 24 | 1 | 1 | 1 | 1 | 1 | 1 | 1 |

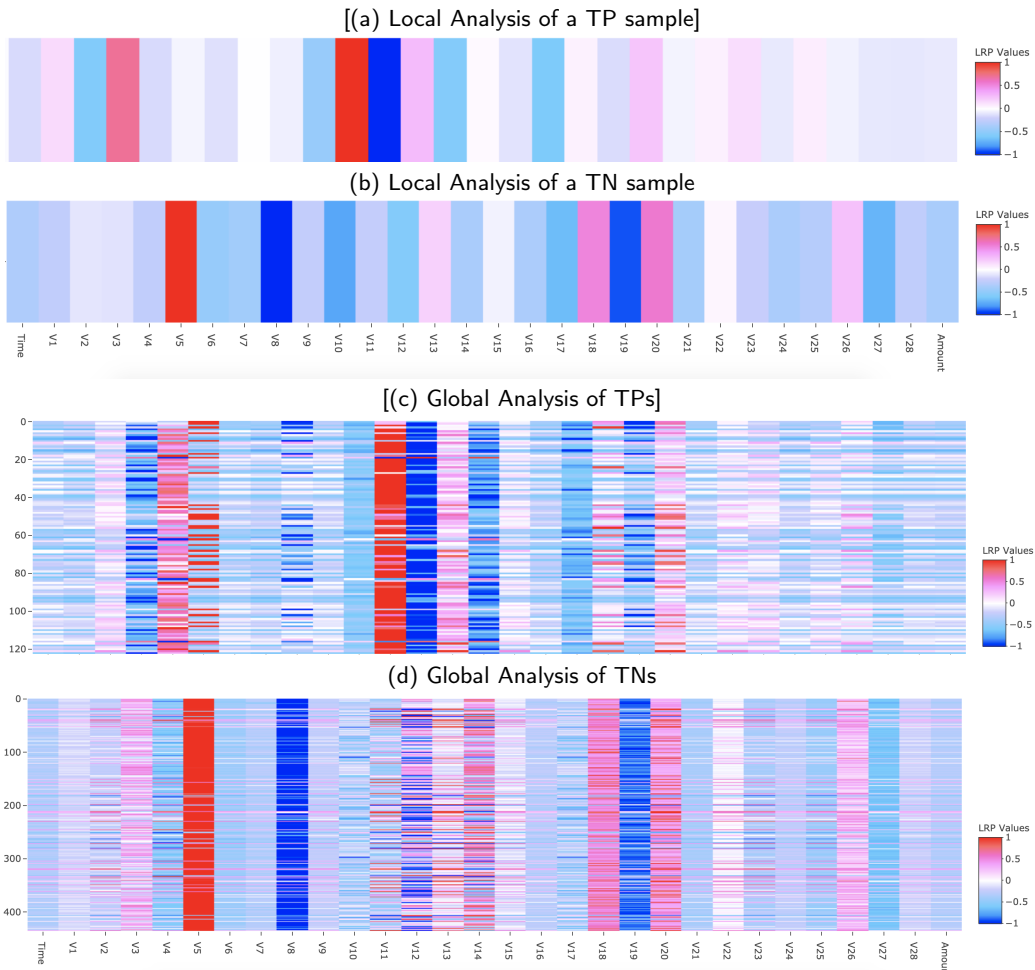

**Figure 5:** Visualizing LRP heatmaps for Local (individual) and global (all) TP and TN samples in CCFDD testing set. Feature sequence of (a) and (c) is similar to (b) and (d)

**Table 6**
Ranked feature comparison for LRP, SHAP, and LIME over creditcard dataset

|  | V11 | V4 | V5 | V20 | V13 | V18 | V14 | V12 | V3 | V26 | V22 | V2 | V6 | V7 | V8 | V9 | V10 | V1 | Amount | V28 | V15 | V16 | V17 | V19 | V21 | V23 | V24 | V25 | V27 | Time |
|---|---|---|---|---|---|---|---|---|---|---|---|---|---|---|---|---|---|---|---|---|---|---|---|---|---|---|---|---|---|---|
| LRP | 30 | 29 | 28 | 27 | 26 | 25 | 24 | 23 | 21 | 21 | 20 | 1 | 1 | 1 | 1 | 1 | 1 | 1 | 1 | 1 | 1 | 1 | 1 | 1 | 1 | 1 | 1 | 1 | 1 | 1 |
| LIME | 30 | 25 | 1 | 1 | 23 | 1 | 27 | 29 | 24 | 1 | 1 | 1 | 1 | 1 | 1 | 1 | 28 | 1 | 1 | 1 | 1 | 1 | 25 | 1 | 1 | 1 | 1 | 1 | 1 | 1 |
| SHAP | 28 | 27 | 1 | 1 | 25 | 1 | 28 | 30 | 24 | 1 | 1 | 1 | 1 | 1 | 1 | 1 | 23 | 1 | 1 | 1 | 1 | 1 | 26 | 1 | 1 | 1 | 1 | 1 | 1 | 1 |